\documentclass{article}

\usepackage[nonatbib,preprint]{neurips_2024}
\usepackage{graphicx}

\usepackage[utf8]{inputenc} 
\usepackage[T1]{fontenc}    
\usepackage{hyperref}       
\usepackage{url}            
\usepackage{booktabs}       
\usepackage{amsfonts}       
\usepackage{nicefrac}       
\usepackage{microtype}      
\usepackage{xcolor}         

\title{Modeling Beyond MOS: Quality Assessment Models Must Integrate Context, Reasoning, and Multimodality}

\author{%
  Mohamed Amine KERKOURI\\
  R\&D Lab\\
  F-Initiatives\\
  Paris, France \\
  \texttt{m.a.kerkouri@f-initiatives.com} \\
  \And
  Marouane Tliba \\
  Université d'Orleans \\
  Orleans, France \\
  \texttt{marouane.tliba@univ-orleans.fr} \\
  \AND
  Aladine Chetouani \\
  Université Sorbonne Paris Nord \\
  Paris, France \\
  \texttt{aladine.chetouani@univ-paris13.fr} \\
  \And
  Nour Aburaed \\
  University of Dubai \\
  Dubai, UAE  \\
  \texttt{noaburaed@ud.ac.ae} \\
  \And
  Alessandro Bruno \\
  IULM University \\
  Milan, Italy \\
  \texttt{alessandro.bruno@iulm.it} \\
}

\begin{document}

\maketitle

\begin{abstract}
This position paper argues that Mean Opinion Score (MOS), while historically foundational, is no longer sufficient as the sole supervisory signal for multimedia quality assessment models. MOS reduces rich, context-sensitive human judgments to a single scalar, obscuring semantic failures, user intent, and the rationale behind quality decisions. We contend that modern quality assessment models must integrate three interdependent capabilities: (1) context-awareness, to adapt evaluations to task-specific goals and viewing conditions; (2) reasoning, to produce interpretable, evidence-grounded justifications for quality judgments; and (3) multimodality, to align perceptual and semantic cues using vision–language models. We critique the limitations of current MOS-centric benchmarks and propose a roadmap for reform: richer datasets with contextual metadata and expert rationales, and new evaluation metrics that assess semantic alignment, reasoning fidelity, and contextual sensitivity. By reframing quality assessment as a contextual, explainable, and multimodal modeling task, we aim to catalyze a shift toward more robust, human-aligned, and trustworthy evaluation systems.
\end{abstract}

\section{Introduction}

Multimedia Quality Assessment (MQA) \cite{multimedia_survey} spans a wide range of tasks, including Image Quality Assessment (IQA) \cite{image_survey}, Video Quality Assessment (VQA) \cite{video_survey}, Audio Quality Assessment (AQA) \cite{multimedia_survey}, and Aesthetic Quality Assessment \cite{aesth_survey}. These models are foundational to applications such as video streaming \cite{straming_survey,ghasempour2025real}, teleconferencing \cite{ying2022telepresence,fadzli2023systematic}, and social media delivery \cite{CAI2024e29020}, where perceptual quality must be optimized under bandwidth and hardware constraints. But they are equally critical in high-stakes domains such as autonomous driving \cite{geng2022robust}, medical imaging \cite{unsupervised_tliba_medical}, and immersive AR/VR environments \cite{tliba_self_supervised_3D}, where even subtle degradations can compromise safety, diagnostic accuracy, or user immersion.

Despite the diversity and complexity of these applications, most MQA systems remain anchored to a single scalar: the Mean Opinion Score (MOS) \cite{ITURecP910}. Introduced for its simplicity and ease of deployment, MOS has become the default supervisory signal for training and evaluating quality assessment models \cite{image_survey,MOS}. It is deeply embedded in dataset design, benchmarking protocols, and learning objectives across IQA, VQA, and related fields.

\textbf{This paper takes the position that MOS, while historically foundational, is no longer sufficient to guide the development of modern quality assessment models.} MOS flattens rich, context-sensitive human judgments into a single number. It cannot explain why a piece of content is judged as high or low quality, nor can it adapt to the specific goals, environments, or user intents that define real-world use cases. This limitation is especially acute in no-reference (NR) settings, where models must assess quality without access to a pristine reference  and where semantic failures, contextual mismatches, or task-specific degradations often go undetected.

Consider a few examples: a novice Twitch streamer misconfigures their camera, resulting in poor framing and lighting; a radiological scan contains subtle artifacts that could mislead diagnosis; a generative image appears sharp but contains hallucinated text or broken structure. In each case, a scalar MOS score is not only uninformative it is misleading. What is needed is a modeling framework that reflects how humans actually perceive, reason about, and act on quality.

We argue that quality assessment models must move beyond scalar prediction and integrate three interdependent capabilities:
\begin{itemize}
    \item \textbf{Context-awareness}: to adapt evaluations to specific tasks, user goals, and environmental conditions.
    \item \textbf{Reasoning}: to produce interpretable, evidence-grounded justifications for quality judgments.
    \item \textbf{Multimodality}: to align perceptual and semantic cues using vision--language models and other cross-modal signals.
\end{itemize}

These pillars are not optional enhancements they are necessary foundations for building quality assessment systems that are robust, trustworthy, and aligned with real-world needs. By reframing quality assessment as a contextual, explainable, and multimodal modeling task, we aim to catalyze a shift in how the field defines, supervises, and evaluates perceptual quality.

The remainder of this paper is structured as follows: Section~\ref{sec:background} provides historical context and motivation. Section~\ref{sec:limitations} critiques the limitations of MOS and correlation-based evaluation. Section~\ref{sec:paradigm} introduces our proposed modeling paradigm. Section~\ref{sec:reform} outlines a roadmap for benchmark and metric reform. Section~\ref{sec:alternatives} addresses alternative viewpoints, and Section~\ref{sec:conclusion} concludes with a call to action.

\section{Background and Motivation}
\label{sec:background}

The foundations of multimedia quality assessment (MQA) lie in psychophysical research from the 19th century, where early work by Fechner and others introduced methods to quantify human perception of sensory stimuli such as brightness, contrast, and sharpness \cite{hawkins2011william,burningham1999brief}. Concepts like just-noticeable differences (JNDs) \cite{stern2010JND}, rating scales, and paired comparisons \cite{perez2019pairwise} laid the groundwork for structured perceptual evaluation. Thurstone’s law of comparative judgment \cite{thurstone2017law} formalized the derivation of interval scales from subjective comparisons, providing a theoretical basis for perceptual modeling.

As multimedia systems evolved, the need for standardized evaluation protocols became critical. Organizations such as the EBU, ITU, and VQEG introduced formal subjective testing methodologies. Standards like ITU-R BT.500 \cite{ITURec500} and ITU-T P.910 \cite{ITURecP910} institutionalized the Mean Opinion Score (MOS), a scalar average of human ratings, as the dominant proxy for perceived quality \cite{mohammadi2014subjective}. MOS has since become the default supervisory signal for computational models and the foundation of most benchmark datasets.

While MOS has enabled decades of progress, it imposes a reductive view of human judgment. It collapses rich, context-sensitive evaluations into a single scalar, discarding semantic nuance, decision rationale, and task relevance. Moreover, subjective testing is resource-intensive, requiring controlled environments, large participant pools, and significant time and cost. These limitations have driven the development of automated quality assessment models, typically categorized by their reliance on reference content:
\begin{itemize}
    \item \textbf{Full-Reference (FR)}: Models that compare a distorted signal to a pristine reference (e.g., PSNR \cite{Kotevski2010}, SSIM \cite{SSIM}, VMAF \cite{VMAF}).
    \item \textbf{Reduced-Reference (RR)}: Models that use partial information from the reference \cite{RR1,RR2}.
    \item \textbf{No-Reference (NR)}: Models that operate without any reference, relying solely on the distorted input (e.g., BRISQE \cite{BRISQE}, NIQE \cite{niqe}, ARNIQA \cite{ARNIQA}).
\end{itemize}

Deep learning has significantly advanced MQA. CNN-based models such as DB-CNN \cite{dbcnn}, KonCept512 \cite{koniq}, LPIPS \cite{LPIPS}, and TPOIQ \cite{TOPIQ}, and transformer-based architectures like MUSIQ \cite{MUSIQ}, Chen et al. \cite{Cheon_2021_CVPR}, and Re-IQA \cite{saha2023re} have achieved strong performance in both FR and NR settings. However, these models are almost universally trained to regress MOS values. This scalar supervision constrains their expressiveness, pushing them to optimize for perceptual fidelity while suppressing their capacity to model semantic coherence, contextual relevance, or user intent, even when their architectures are capable of doing so.

In these models, context is not explicitly modeled, it is inherited from the subjective testing protocol of the training dataset. A model trained on a single dataset becomes implicitly conditioned on that dataset’s assumptions (e.g., display type, viewing distance, task). When trained on multiple datasets, the model effectively averages across contexts, diluting its ability to adapt to any specific use case. This leads to brittle generalization and a lack of task-awareness in deployment.

Recent work has explored vision--language models (VLMs) such as CLIP \cite{CLIP} and BLIP \cite{BLIP}, which jointly embed visual and textual information and offer a path toward semantically grounded quality assessment \cite{li2025benchmark}. Extensions like CLIP-IQA \cite{wang2022exploring}, QualiCLIP  \cite{agnolucci2024quality}, and Q-Align \cite{wu2023q} adapt these models to perceptual quality tasks. However, they are still trained on the same MOS-annotated datasets as traditional models, and thus inherit the same contextual limitations. Moreover, they lack structured reasoning capabilities, which are essential for producing interpretable and task-aware assessments.

Multimodal foundation models such as Flamingo \cite{Flamingo}, LLaVA \cite{Llava}, and Qwen-VL \cite{Qwen-VL} further expand the potential of quality assessment by integrating vision-language alignment with instruction-following capabilities. These systems can generate natural language rationales, interpret visual content in context, and perform multi-step reasoning. However, recent evaluations \cite{wu2024comprehensive} show that even state-of-the-art models like GPT-4V struggle to detect fine-grained perceptual artifacts and align with human quality judgments, despite their massive scale.

A particularly promising direction is the emergence of Chain-of-Thought (CoT) prompting and reinforcement learning (RL) fine-tuning \cite{deepseekai2025r1,qwq32b}, which enable models to articulate step-by-step reasoning processes. These techniques have improved performance on tasks requiring multi-step inference and justification, and are increasingly being explored in vision-language contexts \cite{shen2025vlm}. RL-based strategies such as GRPO \cite{GRPO} and PPO \cite{PPO} further enhance alignment with human-like reasoning and preferences.

Finally, we note that most current models underutilize other perceptually and cognitively relevant modalities such as visual attention maps, saliency cues, and artifact localization masks, which could serve as valuable signals for grounding quality judgments in both perceptual and semantic evidence. These modalities remain largely disconnected from current modeling pipelines and are rarely integrated with language-based reasoning or contextual conditioning.

Together, these developments underscore both the feasibility and the urgency of a paradigm shift. \textbf{We argue that multimedia quality assessment must move beyond scalar MOS prediction and embrace a modeling framework that is multimodal, explainable, and context-aware one that reflects the richness of human judgment and the complexity of modern multimedia content.}

\section{The Limits of MOS and Correlation-Based Evaluation}
\label{sec:limitations}

While MOS remains the de facto standard for image/video quality benchmarks, its single‑scalar label \textbf{actively hinders} modern systems. By collapsing human judgments into one number, MOS masks semantic failures, erases context, and trains models to chase averages instead of edge‑case critical errors. Below, we explore four fundamental flaws of MOS‑based training and evaluation.

\subsection{MOS as a Bottleneck for Generalization and Semantics}

MOS supervision inherently restricts model outputs to a scalar regression target, eliminating the capacity to encode which attributes drive quality assessments. Liu and Bovik \cite{5756237} demonstrated that IQA models trained exclusively on MOS labels routinely misclassify semantically broken artifacts (e.g., hallucinated text) as high quality, despite appropriately down‑weighting benign degradations (e.g., mild blur). Recent transformer‑based and vision–language scorers achieve competitive MOS prediction accuracy yet fail to preserve interpretability or object‑level reasoning \cite{wu2023q}.

Also, all contextual metadata like display type, viewing distance, ambient lighting, intended downstream task is baked into the original subjective test protocol (e.g.\ ITU‑T P.910) rather than exposed to the model as features \cite{ITURecP910}. Consequently, a model trained on a single MOS dataset implicitly adopts its lab’s assumptions. When you train across multiple datasets, the model effectively averages over conflicting contexts: in \cite{9022237}, cross‑dataset SROCC drops by 25\% on no‑reference benchmarks.

In no‑reference (NR) settings, the absence of any pristine reference further amplifies this limitation: models must infer quality solely from semantic cues, which hinders their ability to distinguish task‑critical distortions from benign variations. Zhu et al. \cite{zhu2025semantically} report that conventional NR metrics such as VMAF \cite{VMAF} systematically underrate visually sharp but semantically empty gameplay frames, whereas a CLIP‑guided NR model that dynamically reweights feature contributions based on scene context yields a 0.15 increase in Spearman’s correlation on an out‑of‑domain “in‑the‑wild” games dataset. These findings underscore that MOS‑based supervision does not encode content relevance, unless implicitly specified during data collection.





\subsection{Correlation Does Not Imply Understanding}

Contemporary quality‑assessment benchmarks judge models almost exclusively by their Pearson Linear Correlation Coefficient (PLCC), Spearman Rank Order Correlation Coefficient (SROCC), or Root Mean Squared Error (RMSE) with respect to averaged MOS labels. These metrics quantify statistical alignment with mean human scores but provide no guarantee that a model captures the mechanisms of human perception or the semantics underlying quality judgments \cite{kastryulin2022pytorch}, \cite{hammou2025image}. High correlation thus reflects only that a model has learned to match dataset‑wide averages, not that it truly “understands” why one distortion is more objectionable than another \cite{ma2025survey}.
Empirical cross‑dataset evaluations reveal this gap starkly. Saha et al. (Re‑IQA) demonstrate that leading no‑reference IQA methods, achieving SROCC > 0.95 on synthetic distortions suffer a \(~30\%\) drop in SROCC when tested on authentic “in‑the‑wild” images (e.g., KonIQ‑10k, CLIVE) \cite{saha2023re}. 
Yang et al. \cite{9022237}further report clear performance degradation of CNN‑based NR‑IQA models in cross‑dataset settings, despite strong in‑domain correlations. 
Hosu et al. confirm that models with  \(PLCC\simeq0.92\)  on curated datasets fall below \(SROCC = 0.70\) on KonIQ‑10k’s authentic distortions 
\cite{koniq10k}. 
In the video domain, Li et al. \cite{li2019quality} show that VQA algorithms tuned for lab‑style degradations fail to generalize to in‑the‑wild shooting conditions, incurring substantial correlation losses. Ghadiyaram \& Bovik \cite{ghadiyaram2017perceptual} likewise find that models trained on synthetic distortions poorly predict quality on real‑world mixtures of distortions.

Beyond cross‑domain brittleness, correlation‑driven training inherently biases models toward the dataset mean, suppressing uncertainty and down‑weighting rare but critical distortions. As a result, models optimized for PLCC / SROCC miss high-impact artifacts, semantic anomalies, edge case degradations, and context-sensitive errors, because they are in the tails of the MOS distribution \cite{dong2025exploring}. Recent works like STNS‑IQA demonstrate that incorporating scene‑aware statistical features and multimodal cues (e.g., CLIP embeddings) recovers over 0.20 SROCC in GAN‑generated images where baseline NR metrics fail \cite{yang2024no}. Similarly, IE‑IQA integrates intelligibility features to improve robustness across contexts, highlighting the necessity of moving beyond pure correlation objectives 
\cite{song2021ie}.

These findings confirm that high correlation with MOS benchmarks does not imply human‑aligned understanding or robust perceptual fidelity. We therefore advocate for augmenting existing evaluation protocols with :\begin{enumerate}
    \item \textbf{Semantic stress tests} that probe object, and scene level failure.
    \item \textbf{Uncertainty calibration metrics} to reward explicit confidence modeling
    \item \textbf{Context‑aware benchmarks} that expose models to a diversity of viewing conditions and user intents
\end{enumerate} Only by expanding beyond scalar correlation can we ensure quality‑assessment systems that genuinely reflect human judgments.

\subsection{Structural Failures and the Collapse of Disagreement}

MOS reduces inter‑subject and intra‑subject variability signals of perceptual ambiguity and cultural diversity into a single scalar, thereby discarding information about rating spread and consensus strength \cite{ghadiyaram2015massive}. In the KonIQ‑10k dataset \cite{koniq10k}, per‑image rating standard deviations range from 0.6 to 1.5 on a five‑point scale, correlating with semantic complexity and demonstrating that images with identical MOS can evoke vastly different agreements among observers. The BIQ2021 database similarly reports that images sharing the same mean score can exhibit variance differences exceeding 1.0, indicating fundamentally distinct perceptual gray zones that a single MOS cannot capture \cite{ahmed2022biq2021}.

Large‑scale crowdsourced experiments on the LIVE‑itW database reveal inter‑subject variance up to 20 points on a 100‑point scale under uncontrolled viewing conditions, yet MOS aggregation obscures these environmental sensitivities \cite{babu2023no}. By averaging over these distributions, MOS‑trained models become oblivious to borderline or divisive content and cannot flag stimuli that straddle acceptability thresholds for different user groups \cite{ying2021patch}. The Konstanz Natural Video Database (KoNViD‑1k)\cite{hosu2017konstanz} shows that videos with similar MOS can have rating variances driven by motion ambiguity, yet standard VQA models treat them as uniform quality. This collapse of disagreement limits personalization and accessibility, as models lack the capacity to adjust to individual or cultural preferences.

Furthermore, MOS‑centric training underrepresents rare but critical artifacts: generative‑image distortions such as semantic misalignments and text hallucinations are virtually absent in legacy datasets, causing no‑reference IQA models to assign high quality to AI‑generated images despite glaring semantic errors somerecent works like \cite{li2023agiqa} are trying to adress this gap. Similar blind spots occur for short‑form video distortions: e.g. frame freezing and ghosting in KVQ, where novel artifacts provoke high disagreement but remain unnoticed when averaged to MOS \cite{lu2024kvq}. Multi‑angle video assessments also reveal that wide‑angle distortions yield high inter‑subject variance, yet MOS aggregation masks these differences \cite{hu2025multi}. Even in laboratory‑controlled NITS‑IQA experiments, significant rating spreads persist under uniform conditions, underscoring that disagreement is not noise but essential information \cite{ruikar2023nits}.

Addressing this collapse requires preserving and modeling rating distributions via histograms, confidence intervals, or uncertainty estimates, rather than reducing them to a single mean. Only by capturing disagreement can quality assessment models identify divisive content, support personalization, and remain robust to novel, context‑sensitive artifacts.





\subsection{MOS Discourages Interpretability and Explainability}

Mean Opinion Score supervision provides only a scalar target, offering neither the rationale behind a quality judgment nor any spatial or semantic localization of artifacts, especially in NR settings. As a result, MOS‑trained models function as black boxes: there is no mechanism to trace \textit{why} a given image or video frame is assigned a specific score. This opacity is untenable in high‑stakes domains such as medical imaging or autonomous driving, where understanding model decisions is essential for trust and safety.

Explainable IQA methods illustrate the alternative: Kazemi Ranjbar and Fatemizadeh’s ExIQA framework \cite{kazemi2024exiqa} predicts distortion types and strengths using vision–language models and outputs both a quality score and a distortion attribution map. Similarly, recent work on semantic‑attribute reasoning in BIQA not only estimates MOS but also produces interpretable feature attributions that highlight object‑ or scene‑level factors influencing quality \cite{huang2022explainable}. In medical imaging, specialized explainability approaches generate high‑resolution saliency maps that reveal which anatomical regions drive the quality assessment, enabling radiologists to verify and correct model outputs \cite{amanova2022explainability}.

By contrast, MOS‑only models cannot support human‑in‑the‑loop workflows, adaptive enhancement, or actionable feedback capabilities demonstrated by explainable IQA systems. Without structured explanations, developers cannot debug failure modes, regulators cannot audit compliance, and end users cannot gain confidence in model reliability. We therefore argue that future supervisory signals must integrate interpretability objectives forcing models to justify their scores with localized, evidence‑grounded explanations.  Recent emerging reasoning CoT in reasoning models like DeepSeek-R1 \cite{deepseekai2025r1} could also generate a type of interpretability that is more organic to human understanding instead of the post-hoc mechanistic explainability methods used in other approaches.




\subsection{MOS Trains Models to Be Perceptually Myopic}

Models trained solely to minimize MOS regression losses disproportionately optimize for statistically frequent, low‐level distortions blur, noise, blocking while remaining blind to rare distortions, semantically catastrophic or contextually unacceptable anomalies. For instance, analyses on the PIPAL\cite{jinjin2020pipal} dataset reveal that state‑of‑the‑art IQA metrics (e.g.\ SSIM, LPIPS) achieve only moderate Spearman correlations on GAN‑generated outputs, failing to penalize semantic misalignments that humans deem unacceptable. Similarly, MANIQA demonstrates that NR‑IQA networks must be explicitly tailored to GAN distortions to recover just a fraction of their performance on synthetic benchmarks \cite{yang2022maniqa}.

Hallucinated or implausible content, such as extra limbs or nonsensical text in generative images, can nonetheless receive high scores from MOS‑based predictors. The AGIQA‑3K  \cite{li2023agiqa} study shows that off‑the‑shelf NR‑IQA models correlate poorly with human judgments of text‑to‑image alignment and semantic fidelity, whereas dedicated semantic metrics substantially improve assessment on text‑guided generative outputs. Large multimodal models further confirm this gap: conventional DNN‑based IQA methods underperform on AI‑generated images with semantic errors, while CLIP‑guided architectures regain over 0.20 SROCC by incorporating semantic embeddings \cite{wang2024large}.

In the video domain, MOS‑centric VQA also neglects temporal artifacts and rare but perceptually salient events. The PEA265 database documents six common compression artifacts, including flickering and floating—that standard VQA algorithms fail to detect without bespoke temporal modeling \cite{lin2020pea265}. Saliency‑Aware Spatio‑Temporal Artifact Measurement (SSTAM) further shows that no‑reference VQA methods overlook local flicker unless explicitly trained on artifact annotations, resulting in large drops in correlation with MOS on flicker‑rich sequences \cite{lin2023saliency}. 

These patterns of failure underscore that MOS training instills a form of perceptual myopia: models learn to reward smoothness over structure, sharpness over meaning, and frequency over significance. Emerging approaches, such as IE‑IQA’s integration of intelligibility features \cite{song2021ie}, CLIP‑AGIQA’s exploit of visual–textual alignment \cite{tang2025clip}, and prompt‑aware IQA frameworks \cite{qu2024bringing} demonstrate the necessity of multimodal, semantics‑driven supervision. To overcome perceptual myopia, future benchmarks must incorporate semantic anomalies, context‑aware distortions, and uncertainty modeling rather than rely exclusively on MOS correlation.

\section{Modeling Beyond MOS: Toward Contextual, Explainable, and Multimodal Quality Assessment}
\label{sec:paradigm}

\begin{figure}
    \centering
    \includegraphics[width=1\linewidth]{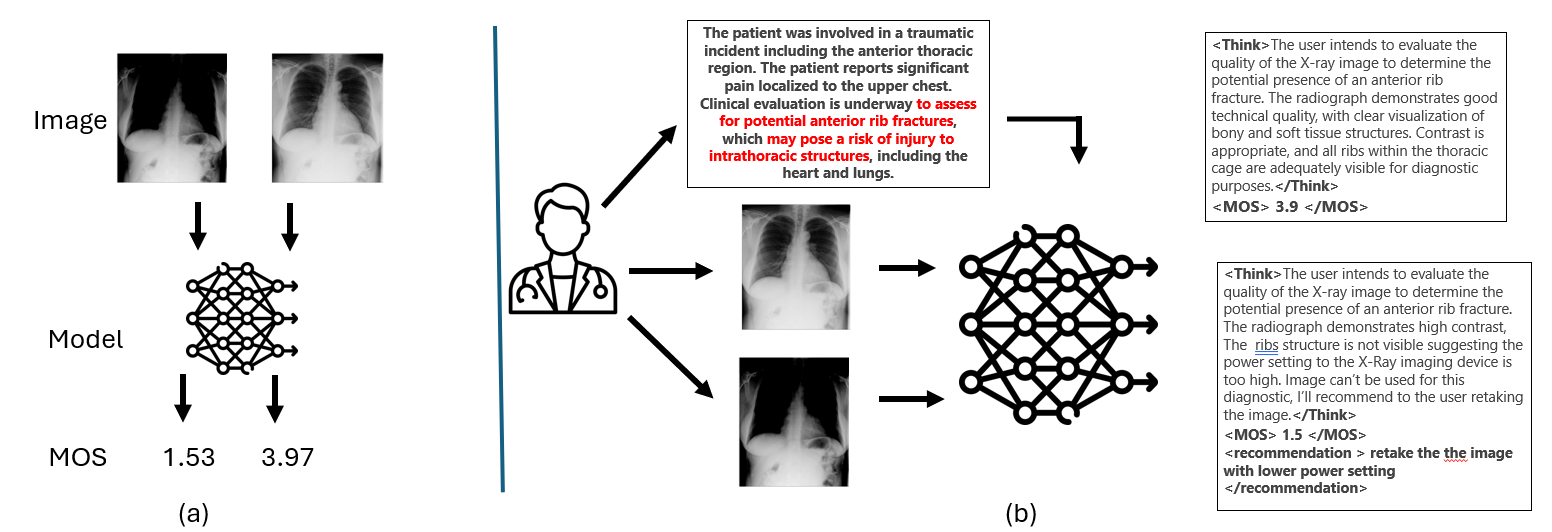}
    \caption{An illustrative example comparing the a reduced pipeline from our paradigm (b) with a traditional MOS quality assessment system (a) in a clinical setting. By providing the context to the model, it can focus on estimating the quality for a specific task, as well as give recommendations, all while the reasoning behind the decision and recommendation stay transparent through the thinking CoT.}
    \label{fig:enter-label}
\end{figure}

To overcome the structural limitations of MOS-based supervision, we argue for a modeling paradigm that is explicitly \textbf{context-aware}, \textbf{reasoning-capable}, and \textbf{multimodal}. These are not optional enhancements, they are necessary foundations for building quality assessment systems that are robust, interpretable, and aligned with real-world use cases beyond streaming and compression. In this section, we outline how each of these pillars addresses specific failure modes identified in Section~\ref{sec:limitations}, and how they can be operationalized in model design.

\subsection{Context-Aware Modeling}

MOS‑trained models inherit context implicitly from the protocols under which subjective scores were collected, but they lack mechanisms to reason about task, user intent, or viewing conditions \cite{ITURecP910}. A truly context‑aware model must accept explicit metadata inputs such as task type (e.g., medical diagnosis vs. social streaming) or device parameters (e.g., mobile display vs. VR headset) and condition its quality predictions accordingly \cite{zhu2020metaiqa}. MetaIQA employs meta‑learning to adapt NR‑IQA models to unknown distortions by leveraging task‑level priors, demonstrating substantial cross‑domain gains when context shifts \cite{zhu2020metaiqa}.  Semantic‑guided NR‑IQA (SGIQA) integrates scene‑category prompts into a Swin Transformer backbone, yielding a 0.07 SROCC lift over context‑agnostic counterparts on in‑the‑wild datasets \cite{pan2024sgiqa}.

Context-awareness enables models to distinguish between acceptable and unacceptable degradations based on use case. A compression artifact that is tolerable in a social media clip may be unacceptable in a radiological scan. Without explicit context, models collapse these distinctions, a failure mode that cannot be corrected by MOS regression alone.

\subsection{Reasoning-Centric Quality Assessment}

MOS supervision discourages interpretability. To reverse this, we advocate for models that generate \textbf{explicit chains of reasoning}  structured, step-by-step justifications for their quality judgments. This can be achieved through Chain-of-Thought (CoT) prompting, vision-language entailment modules, or reinforcement learning (RL) fine-tuning with reasoning-aligned objectives.

Each reasoning step should be grounded in visual evidence and validated for factual consistency. For example, if a model claims “the license plate is unreadable,” it must attend to the relevant region and verify that the claim is supported by the image. This closes the loop between \textit{where} the model looks and \textit{why} it makes a decision. Reasoning also enables models to expose internal conflicts  such as high technical fidelity but low semantic integrity, which scalar scores obscure.

Recent advances in multimodal CoT generation \cite{deepseekai2025r1} and RL-based alignment \cite{GRPO} demonstrate that such reasoning is not only feasible but improves robustness and human trust. In quality assessment, it enables models to surface actionable insights, support human-in-the-loop workflows, and justify decisions in high-stakes domains.

\subsection{Multimodal Fusion and Alignment}

Human quality judgments are inherently multimodal: they arise from the interplay of visual perception, semantic understanding, and contextual intent. Yet most models reduce this process to a single perceptual embedding. We argue that quality assessment must integrate at least four complementary modalities:
\begin{enumerate}
    \item \textbf{Visual features}: raw pixels or deep perceptual embeddings.
    \item \textbf{Textual context}: task descriptions, user roles, or usage scenarios.
    \item \textbf{Visual attention cues}: saliency maps, gaze data, or learned attention heatmaps , \cite{kerkouri2022domain}, \cite{tliba2022self}, \cite{chetouani2020use}.
    \item \textbf{Artifact maps}: localized distortion masks or hallucination flags.
\end{enumerate}

These modalities should be embedded into a joint representation space using a vision-language backbone augmented with attention-fusion layers. This enables a series of alignment checks:
\begin{itemize}
    \item \textbf{Image-attention alignment}: ensures that high-salience regions coincide with perceptually degraded or semantically critical areas.
    \item \textbf{Attention-text alignment}: verifies that textual rationales refer to regions that attracted attention.
    \item \textbf{Artifact-text alignment}: confirms that semantic failure flags correspond to actual distortion masks.
\end{itemize}

Such alignment is not a luxury, it is a prerequisite for trustworthy, human-aligned quality assessment. Without it, models hallucinate explanations, ignore critical regions, or reward perceptual smoothness over semantic fidelity.

\subsection{From Scalar Prediction to Structured Judgment}

Together, these modeling principles redefine quality assessment as a structured prediction task instead of a scalar regression problem. A model should output:
\begin{itemize}
    \item A contextualized quality score (conditioned on task and user intent).
    \item A natural language rationale (explaining the judgment).
    \item A visual attention map (showing where the model looked).
    \item An artifact mask (highlighting localized distortions).
\end{itemize}

This structured output enables richer supervision, more informative evaluation, and actionable feedback. It also aligns with emerging trends in multimodal AI, where interpretability, grounding, and task adaptation are no longer optional.

\section{Reforming Benchmarks for Contextual and Explainable Quality Assessment}
\label{sec:reform}

To align evaluation with the context‐ and explainability‐driven paradigm outlined in Section \ref{sec:paradigm}, benchmarks must shed their dependence on legacy lab‑based MOS scores and instead integrate explicit task metadata, persona‑conditioned judgments, and rationale annotations. We next detail four targeted reforms, spanning data collection, annotation protocols, simulation, and multi‑facet metrics—that realize this shift without expanding benchmark size.


\subsection{Contextual and Multi‑Perspective Annotation Protocols}

To capture the full complexity of real‑world quality judgments, benchmarks must go beyond single‑condition Opinion Score ratings and incorporate diverse perspectives and explanatory signals:

\begin{itemize}
  \item \textbf{Multi‑condition labeling}: Collect separate quality scores for each predefined scenario (e.g., “clinical diagnosis on MRI,” “social media upload on smartphone,” “immersive viewing in VR”), with metadata for task, device, and environment, and explanations to subject before experience.  
  \item \textbf{Role‑specific ratings}: Assign annotators explicit personas (e.g., radiologist, video editor, end‑user) so that perceptual criteria and resulting label distributions, reflect varied expertise and priorities, this can go even further through emotion detection by facial expression\cite{singh16emotion} and micro-expression \cite{jha2025human}.  
  \item \textbf{Structured rationales}: Require a concise justification for each score, such as a chain‑of‑thought outline or attribute checklist (“text legibility,” “artifact severity”)—on a representative subset of items.  
  \item \textbf{Semantic‑error flags}: Beyond quality scores, annotators must mark semantic failures (e.g., hallucinated objects, missing critical details) that MOS alone cannot expose.  
  \item \textbf{Distributional labels}: Preserve the full histogram or confidence interval of ratings per condition and persona rather than a single mean, enabling models to learn from disagreement and quantify uncertainty,as well rare cases.  
\end{itemize}

\subsection{Scalable and Targeted Data Collection Infrastructure}

Deploying the annotation protocols outlined above at scale necessitates specialized infrastructure for diverse, high‑quality labels:

\begin{itemize}
  \item \textbf{Contextual crowdsourcing platforms}: Integrate scenario scripts and persona prompts into task interfaces, ensuring raters understand the specific viewing conditions and user roles before annotation \cite{xu2021subjective}.  

  \item \textbf{In‑the‑wild mobile annotation apps}: Embed rating, rationale entry, and declared intent (e.g., “uploading to Instagram,” “diagnosing an X‑ray”) into consumer‑facing apps, capturing authentic user judgments in situ.  

  \item \textbf{Active learning pipelines}: Use preliminary models to identify semantic edge cases or images with high predicted quality but low MOS, and prioritize these for expert re‑annotation, thereby improving label efficiency by up to 40\% \cite{li2024survey}.  

  \item \textbf{Gamified annotation interfaces}: Employ game mechanics, such as role‑based challenges (“spot the compression artifact as a radiologist”) and point systems—to boost annotator engagement and depth of rationale, which has been shown to increase both speed and accuracy in image description tasks \cite{ivanjko2019crowdsourcing}.  

  \item \textbf{Longitudinal expert panels}: Convene panels of domain experts to re‑rate a subset of content at regular intervals, revealing perceptual drift and evolving standards; the LEAD methodology achieved inter‑rater reliability > 0.85 across repeated conditions \cite{EIJSBROEK2025152603}.  
\end{itemize}

\subsection{Simulation and Evaluation with Persona-Conditioned Models}

With context- and persona-rich datasets, benchmarks can support simulation-based evaluation:

\begin{itemize}
    \item \textbf{Conditional quality generators}: A reasoning generative model takes an image and a persona-context embedding and outputs an “opinion overlay”, an explanation of perceived flaws or strengths.
    
    \item \textbf{Multi-agent rating simulation}: Multiple persona agents (e.g., radiologist, streamer, casual viewer) each generate a quality score and rationale using their own reasoning policy. This yields a distribution of judgments and explanations \cite{ge2025scalingsyntheticdatacreation}.
    
    \item \textbf{ Reasoning Metrics}: Introducing recent metrics that evaluate coherence, validity, groundness and factuality \cite{fabbri-etal-2022-qafacteval}, and  utility  of reasoning in LLMs and Multimodal LLMs \cite{Lee2025EvaluatingSR} must be used as training and evaluation objectives alongside perceptual losses.
    
    \item \textbf{Few-shot persona adaptation}: A lightweight meta-learner adapts a model’s scoring head to new personas or tasks with minimal supervision, enabling flexible deployment across domains \cite{ge2025scalingsyntheticdatacreation}.
\end{itemize}

\subsection{Toward Scientifically Grounded Evaluation Platforms}

These reforms transform benchmarks from static MOS collections into scientifically grounded, multi-dimensional evaluation platforms. They support:

\begin{itemize}
    \item \textbf{Context-aware evaluation}: Models are tested under varying task and user conditions.
    \item \textbf{Explainability assessment}: Rationales and attention maps are evaluated for coherence and grounding.
    \item \textbf{Semantic robustness}: Models are challenged with hallucinations, rare artifacts, and ambiguous content.
    \item \textbf{Personalization and fairness}: Rating distributions and persona diversity are preserved and modeled.
\end{itemize}

These are not speculative enhancements, they are necessary conditions for evaluating the next generation of quality assessment systems, and some preliminary works are being conducted in some aspects. Without them, we cannot measure what matters: whether a model understands quality, adapts to context, and explains its decisions, and by extension accurately model the human visual system behavior.

\section{Alternative Viewpoints and Rebuttals}
\label{sec:alternatives}

\textbf{“MOS is sufficient.”}  
Mean Opinion Score (MOS) has indeed powered decades of quality assessment, offering a single, interpretable metric collected under standardized protocols (e.g., ITU‑T P.910) \cite{ITURecP910}. Its simplicity enabled rapid dataset creation and straightforward benchmarking\cite{koniq}, \cite{TID2013}. However, this very simplicity glosses over context‐dependent semantics, user intent, and the reasoning behind judgments as shown in section \ref{sec:limitations}. We do not discard MOS; rather, we retain it as a baseline while augmenting benchmarks with context‐conditioned scores, semantic‐failure flags, and reasoning metrics. In constrained scenarios, pure MOS suffices, but in complex or safety‐critical domains like medical imaging \cite{unsupervised_tliba_medical} or autonomous driving\cite{10726865}, additional axes are essential to surface failures that MOS alone conceals.

\textbf{“Explainability isn’t necessary.”}  
It is argued that users only need a numerical score, not a chain of thought. Yet extensive work in explainable AI demonstrates that transparency is indispensable when trust and accountability matter \cite{10.1145/3705724}, \cite{Brandenburg2025}, \cite{Balasubramaniam2022}. In medical workflows, radiologists require saliency maps and rationales to validate automated assessments \cite{10.1145/3617233.3617234}. In content‐creation pipelines, interpretable feedback accelerates debugging and user learning \cite{EngelHermann2024}. Without explicit explanations, models remain opaque black boxes, hindering regulatory approval and end‐user confidence.

\textbf{“Computational complexity concerns.”}  
Integrating multimodal reasoning increases system footprint. Yet modern architectures achieve efficient conditional computation, Mixture‐of‐Experts (MoE) layers route only relevant tokens through heavy subnets \cite{Masoudnia2014} leading to substantial savings in computational power exceeding 50\%. Adapter modules permit on‐demand reasoning at only a fraction of full‐model cost \cite{Hu2022}.  We advocate tiered pipelines: a MOS‑only fast path for low‑risk use, and a full reasoning path for high‑stakes tasks.

\textbf{“Subjectivity makes standardization impossible.”}  
Human judgments vary by culture, expertise, and intent, yet fields such as usability and human–computer interaction routinely formalize subjective assessments into repeatable protocols like ISO9241 \footnote{\href{https://www.iso.org/standard/77520.html}{https://www.iso.org/standard/77520.html}}. By explicitly modeling personas and contexts, we embrace rather than erase diversity. Persona‐conditioned benchmarks reveal demographic sensitivities and enable fairness analyses . Multi‐agent evaluation frameworks from dialogue systems illustrate how divergent user profiles can be jointly modeled and simulated \cite{ge2025scalingsyntheticdatacreation}. Rather than a barrier, subjectivity becomes an asset for creating robust, personalized quality‐assessment systems.

\section{Conclusion \& Call to Action}
\label{sec:conclusion}

\textbf{MOS alone can no longer serve as the sole supervisory signal for multimedia quality assessment; it must be complemented by explicit \emph{context‑awareness}, structured \emph{reasoning}, and integrated \emph{multimodality}.} By flattening nuanced, task‑dependent human judgments into a single number, MOS‑centric approaches obscure semantic failures, disregard user intent and viewing conditions, and provide no insight into the reasoning behind quality decisions. In this paper, we have illustrated these shortcomings with concrete examples from previous studies, outlined a blueprint for richer benchmarks including context‑conditioned labels, persona‑driven ratings, rationale annotations, semantic‑failure flags, and rating distributions and proposed novel evaluation metrics that examine “why,” “where,” and “when” model outputs align with human perception.

The time has come to evolve our benchmarks and modeling paradigms. In the near term, we encourage the release of datasets annotated with task and persona metadata, structured rationales, and full rating distributions; the organization of shared challenges that assess semantic robustness, explainability, and context sensitivity alongside traditional correlation metrics; and the incorporation of differentiable reasoning and multimodal alignment objectives into model training. Over the longer horizon, we envision interactive, real‑time quality advisors that adapt to individual user needs, collaborative platforms for crowdsourced rationale collection, and interdisciplinary efforts to define ethical standards for context and persona modeling. Embracing this expanded framework will yield multimedia quality‑assessment systems that are not only accurate but genuinely human‑aligned, transparent, and trustworthy.

\medskip

{\small
\bibliographystyle{IEEEbib}
\bibliography{strings}

\begin{thebibliography}{100}

\bibitem{multimedia_survey}
Zahid Akhtar and Tiago~H Falk,
\newblock ``Audio-visual multimedia quality assessment: A comprehensive survey,''
\newblock {\em IEEE access}, vol. 5, pp. 21090--21117, 2017.

\bibitem{image_survey}
Guangtao Zhai and Xiongkuo Min,
\newblock ``Perceptual image quality assessment: a survey,''
\newblock {\em Science China Information Sciences}, vol. 63, pp. 1--52, 2020.

\bibitem{video_survey}
Xiongkuo Min, Huiyu Duan, Wei Sun, Yucheng Zhu, and Guangtao Zhai,
\newblock ``Perceptual video quality assessment: A survey,''
\newblock {\em Science China Information Sciences}, vol. 67, no. 11, pp. 211301, 2024.

\bibitem{aesth_survey}
Yubin Deng, Chen~Change Loy, and Xiaoou Tang,
\newblock ``Image aesthetic assessment: An experimental survey,''
\newblock {\em IEEE Signal Processing Magazine}, vol. 34, no. 4, pp. 80--106, 2017.

\bibitem{straming_survey}
Wei Zhou, Xiongkuo Min, Hong Li, and Qiuping Jiang,
\newblock ``A brief survey on adaptive video streaming quality assessment,''
\newblock {\em Journal of Visual Communication and Image Representation}, vol. 86, pp. 103526, 2022.

\bibitem{ghasempour2025real}
Mohammad Ghasempour, Hadi Amirpour, and Christian Timmerer,
\newblock ``Real-time quality-and energy-aware bitrate ladder construction for live video streaming,''
\newblock {\em IEEE Journal on Emerging and Selected Topics in Circuits and Systems}, 2025.

\bibitem{ying2022telepresence}
Zhenqiang Ying, Deepti Ghadiyaram, and Alan Bovik,
\newblock ``Telepresence video quality assessment,''
\newblock in {\em European Conference on Computer Vision}. Springer, 2022, pp. 327--347.

\bibitem{fadzli2023systematic}
Fazliaty~Edora Fadzli, Ajune~Wanis Ismail, and Shafina Abd Karim~Ishigaki,
\newblock ``A systematic literature review: Real-time 3d reconstruction method for telepresence system,''
\newblock {\em Plos one}, vol. 18, no. 11, pp. e0287155, 2023.

\bibitem{CAI2024e29020}
Qin-Yu Cai, Jing Tang, Si-Zhe Meng, Yi~Sun, Xia Lan, and Tai-Hang Liu,
\newblock ``Quality assessment of videos on social media platforms related to gestational diabetes mellitus in china: A cross-section study,''
\newblock {\em Heliyon}, vol. 10, no. 7, pp. e29020, 2024.

\bibitem{geng2022robust}
Keke Geng, Ge~Dong, and Wenhan Huang,
\newblock ``Robust dual-modal image quality assessment aware deep learning network for traffic targets detection of autonomous vehicles,''
\newblock {\em Multimedia Tools and Applications}, vol. 81, no. 5, pp. 6801--6826, 2022.

\bibitem{unsupervised_tliba_medical}
Marouane Tliba, Aymen Sekhri, Mohamed~Amine Kerkouri, and Aladine Chetouani,
\newblock ``Deep-based quality assessment of medical images through domain adaptation,''
\newblock in {\em 2022 IEEE International Conference on Image Processing (ICIP)}, 2022, pp. 3692--3696.

\bibitem{tliba_self_supervised_3D}
Marouane Tliba, Aladine Chetouani, Giuseppe Valenzise, and Frederic Dufaux,
\newblock ``Representation learning optimization for 3d point cloud quality assessment without reference,''
\newblock 10 2022, pp. 3702--3706.

\bibitem{ITURecP910}
{International Telecommunication Union Telecommunication Standardization Sector (ITU-T)},
\newblock ``Recommendation p.910: Subjective video quality assessment methods for multimedia applications,'' \url{https://www.itu.int/rec/T-REC-P.910-202310-I/en}, 2023,
\newblock ITU-T Recommendation P.910 (10/23).

\bibitem{MOS}
Robert~C Streijl, Stefan Winkler, and David~S Hands,
\newblock ``Mean opinion score (mos) revisited: methods and applications, limitations and alternatives,''
\newblock {\em Multimedia Systems}, vol. 22, no. 2, pp. 213--227, 2016.

\bibitem{hawkins2011william}
Stephanie~L Hawkins,
\newblock ``William james, gustav fechner, and early psychophysics,''
\newblock {\em Frontiers in physiology}, vol. 2, pp. 68, 2011.

\bibitem{burningham1999brief}
Norman Burningham,
\newblock ``A brief review of the history and application of psychometrics and scaling to image quality assessment,''
\newblock in {\em PICS}, 1999, pp. 169--172.

\bibitem{stern2010JND}
Melissa~K Stern and James~H Johnson,
\newblock ``Just noticeable difference,''
\newblock {\em The corsini encyclopedia of psychology}, pp. 1--2, 2010.

\bibitem{perez2019pairwise}
Maria Perez-Ortiz, Aliaksei Mikhailiuk, Emin Zerman, Vedad Hulusic, Giuseppe Valenzise, and Rafa{\l}~K Mantiuk,
\newblock ``From pairwise comparisons and rating to a unified quality scale,''
\newblock {\em IEEE Transactions on Image Processing}, vol. 29, pp. 1139--1151, 2019.

\bibitem{thurstone2017law}
Louis~L Thurstone,
\newblock ``A law of comparative judgment,''
\newblock in {\em Scaling}, pp. 81--92. Routledge, 2017.

\bibitem{ITURec500}
{International Telecommunication Union Radiocommunication Sector (ITU-R)},
\newblock ``Recommendation bt.500: Methodology for the subjective assessment of the quality of television pictures,'' \url{https://www.itu.int/rec/R-REC-BT.500}, 2000,
\newblock ITU-R Recommendation BT.500-10.

\bibitem{mohammadi2014subjective}
Pedram Mohammadi, Abbas Ebrahimi-Moghadam, and Shahram Shirani,
\newblock ``Subjective and objective quality assessment of image: A survey,''
\newblock {\em arXiv preprint arXiv:1406.7799}, 2014.

\bibitem{Kotevski2010}
Z.~Kotevski and P.~Mitrevski,
\newblock ``Experimental comparison of psnr and ssim metrics for video quality estimation,''
\newblock in {\em ICT Innovations 2009}, Danco Davcev and Juan~M. Gómez, Eds., pp. 205--213. Springer, Berlin, Heidelberg, 2010.

\bibitem{SSIM}
BovikAC WangZhou, HR~Sheikh, et~al.,
\newblock ``Image qualityassessment: From errorvisibilitytostructural similarity,''
\newblock {\em IEEE Transon ImageProcessing}, vol. 13, no. 4, pp. 600, 2004.

\bibitem{VMAF}
Zhi Li, Anne Aaron, Ioannis Katsavounidis, Anush Moorthy, and Megha Manohara,
\newblock ``Toward a practical perceptual video quality metric, 2016,''
\newblock {\em Dostupno na: http://techblog. netflix. com/2016/06/toward-practical-perceptual-video. html [16.8. 2022.]}, 2016.

\bibitem{RR1}
Xudong Lv and Z.~Jane Wang,
\newblock ``Reduced-reference image quality assessment based on perceptual image hashing,''
\newblock {\em 2009 16th IEEE International Conference on Image Processing (ICIP)}, pp. 4361--4364, 2009.

\bibitem{RR2}
Xuanqin Mou, Wufeng Xue, and Lei Zhang,
\newblock ``Reduced reference image quality assessment via sub-image similarity based redundancy measurement,''
\newblock in {\em Electronic imaging}, 2012.

\bibitem{BRISQE}
Anish Mittal, Anush~Krishna Moorthy, and Alan~Conrad Bovik,
\newblock ``No-reference image quality assessment in the spatial domain,''
\newblock {\em IEEE Transactions on Image Processing}, vol. 21, no. 12, pp. 4695--4708, 2012.

\bibitem{niqe}
Anish Mittal, Rajiv Soundararajan, and Alan~C Bovik,
\newblock ``Making a “completely blind” image quality analyzer,''
\newblock {\em IEEE Signal processing letters}, vol. 20, no. 3, pp. 209--212, 2012.

\bibitem{ARNIQA}
Lorenzo Agnolucci, Leonardo Galteri, Marco Bertini, and Alberto Del~Bimbo,
\newblock ``Arniqa: Learning distortion manifold for image quality assessment,''
\newblock in {\em Proceedings of the IEEE/CVF Winter Conference on Applications of Computer Vision}, 2024, pp. 189--198.

\bibitem{dbcnn}
Weixia Zhang, Kede Ma, Jia Yan, Dexiang Deng, and Zhou Wang,
\newblock ``Blind image quality assessment using a deep bilinear convolutional neural network,''
\newblock {\em IEEE Transactions on Circuits and Systems for Video Technology}, vol. 30, pp. 36--47, 2019.

\bibitem{koniq}
Vlad Hosu, Hanhe Lin, Tamas Sziranyi, and Dietmar Saupe,
\newblock ``Koniq-10k: An ecologically valid database for deep learning of blind image quality assessment,''
\newblock {\em IEEE Transactions on Image Processing}, vol. 29, pp. 4041--4056, 2020.

\bibitem{LPIPS}
Richard Zhang, Phillip Isola, Alexei~A Efros, Eli Shechtman, and Oliver Wang,
\newblock ``The unreasonable effectiveness of deep features as a perceptual metric,''
\newblock in {\em Proceedings of the IEEE conference on computer vision and pattern recognition}, 2018, pp. 586--595.

\bibitem{TOPIQ}
Chaofeng Chen, Jiadi Mo, Jingwen Hou, Haoning Wu, Liang Liao, Wenxiu Sun, Qiong Yan, and Weisi Lin,
\newblock ``Topiq: A top-down approach from semantics to distortions for image quality assessment,''
\newblock {\em IEEE Transactions on Image Processing}, 2024.

\bibitem{MUSIQ}
Junjie Ke, Qifei Wang, Yilin Wang, Peyman Milanfar, and Feng Yang,
\newblock ``Musiq: Multi-scale image quality transformer,''
\newblock in {\em Proceedings of the IEEE/CVF international conference on computer vision}, 2021, pp. 5148--5157.

\bibitem{Cheon_2021_CVPR}
Manri Cheon, Sung-Jun Yoon, Byungyeon Kang, and Junwoo Lee,
\newblock ``Perceptual image quality assessment with transformers,''
\newblock in {\em Proceedings of the IEEE/CVF Conference on Computer Vision and Pattern Recognition (CVPR) Workshops}, June 2021, pp. 433--442.

\bibitem{saha2023re}
Avinab Saha, Sandeep Mishra, and Alan~C Bovik,
\newblock ``Re-iqa: Unsupervised learning for image quality assessment in the wild,''
\newblock in {\em Proceedings of the IEEE/CVF conference on computer vision and pattern recognition}, 2023, pp. 5846--5855.

\bibitem{CLIP}
Alec Radford, Jong~Wook Kim, Chris Hallacy, Aditya Ramesh, Gabriel Goh, Sandhini Agarwal, Girish Sastry, Amanda Askell, Pamela Mishkin, Jack Clark, et~al.,
\newblock ``Learning transferable visual models from natural language supervision,''
\newblock in {\em International conference on machine learning}. PmLR, 2021, pp. 8748--8763.

\bibitem{BLIP}
Junnan Li, Dongxu Li, Caiming Xiong, and Steven Hoi,
\newblock ``Blip: Bootstrapping language-image pre-training for unified vision-language understanding and generation,''
\newblock in {\em International conference on machine learning}. PMLR, 2022, pp. 12888--12900.

\bibitem{li2025benchmark}
Zongxia Li, Xiyang Wu, Hongyang Du, Huy Nghiem, and Guangyao Shi,
\newblock ``Benchmark evaluations, applications, and challenges of large vision language models: A survey,''
\newblock {\em arXiv preprint arXiv:2501.02189}, vol. 1, 2025.

\bibitem{wang2022exploring}
Jianyi Wang, Kelvin~CK Chan, and Chen~Change Loy,
\newblock ``Exploring clip for assessing the look and feel of images,''
\newblock in {\em AAAI}, 2023.

\bibitem{agnolucci2024quality}
Lorenzo Agnolucci, Leonardo Galteri, and Marco Bertini,
\newblock ``Quality-aware image-text alignment for real-world image quality assessment,''
\newblock {\em arXiv preprint arXiv:2403.11176}, vol. 5, no. 6, 2024.

\bibitem{wu2023q}
Haoning Wu, Zicheng Zhang, Weixia Zhang, Chaofeng Chen, Liang Liao, Chunyi Li, Yixuan Gao, Annan Wang, Erli Zhang, Wenxiu Sun, et~al.,
\newblock ``Q-align: Teaching lmms for visual scoring via discrete text-defined levels,''
\newblock {\em arXiv preprint arXiv:2312.17090}, 2023.

\bibitem{Flamingo}
Jean-Baptiste Alayrac, Jeff Donahue, Pauline Luc, Antoine Miech, Iain Barr, Yana Hasson, Karel Lenc, Arthur Mensch, Katherine Millican, Malcolm Reynolds, et~al.,
\newblock ``Flamingo: a visual language model for few-shot learning,''
\newblock {\em Advances in neural information processing systems}, vol. 35, pp. 23716--23736, 2022.

\bibitem{Llava}
Haotian Liu, Chunyuan Li, Qingyang Wu, and Yong~Jae Lee,
\newblock ``Visual instruction tuning,''
\newblock {\em Advances in neural information processing systems}, vol. 36, pp. 34892--34916, 2023.

\bibitem{Qwen-VL}
Jinze Bai, Shuai Bai, Shusheng Yang, Shijie Wang, Sinan Tan, Peng Wang, Junyang Lin, Chang Zhou, and Jingren Zhou,
\newblock ``Qwen-vl: A versatile vision-language model for understanding, localization, text reading, and beyond,''
\newblock {\em arXiv preprint arXiv:2308.12966}, 2023.

\bibitem{wu2024comprehensive}
Tianhe Wu, Kede Ma, Jie Liang, Yujiu Yang, and Lei Zhang,
\newblock ``A comprehensive study of multimodal large language models for image quality assessment,''
\newblock in {\em European Conference on Computer Vision}. Springer, 2024, pp. 143--160.

\bibitem{deepseekai2025r1}
DeepSeek-AI,
\newblock ``Deepseek-r1: Incentivizing reasoning capability in llms via reinforcement learning,'' 2025.

\bibitem{qwq32b}
Qwen Team,
\newblock ``Qwq-32b: Embracing the power of reinforcement learning,'' March 2025.

\bibitem{shen2025vlm}
Haozhan Shen, Peng Liu, Jingcheng Li, Chunxin Fang, Yibo Ma, Jiajia Liao, Qiaoli Shen, Zilun Zhang, Kangjia Zhao, Qianqian Zhang, et~al.,
\newblock ``Vlm-r1: A stable and generalizable r1-style large vision-language model,''
\newblock {\em arXiv preprint arXiv:2504.07615}, 2025.

\bibitem{GRPO}
Zhihong Shao, Peiyi Wang, Qihao Zhu, Runxin Xu, Junxiao Song, Xiao Bi, Haowei Zhang, Mingchuan Zhang, YK~Li, Y~Wu, et~al.,
\newblock ``Deepseekmath: Pushing the limits of mathematical reasoning in open language models,''
\newblock {\em arXiv preprint arXiv:2402.03300}, 2024.

\bibitem{PPO}
John Schulman, Filip Wolski, Prafulla Dhariwal, Alec Radford, and Oleg Klimov,
\newblock ``Proximal policy optimization algorithms,''
\newblock {\em arXiv preprint arXiv:1707.06347}, 2017.

\bibitem{5756237}
Anush~Krishna Moorthy and Alan~Conrad Bovik,
\newblock ``Blind image quality assessment: From natural scene statistics to perceptual quality,''
\newblock {\em IEEE Transactions on Image Processing}, vol. 20, no. 12, pp. 3350--3364, 2011.

\bibitem{9022237}
Dan Yang, Veli-Tapani Peltoketo, and Joni-Kristian Kämäräinen,
\newblock ``Cnn-based cross-dataset no-reference image quality assessment,''
\newblock in {\em 2019 IEEE/CVF International Conference on Computer Vision Workshop (ICCVW)}, 2019, pp. 3913--3921.

\bibitem{zhu2025semantically}
Kai Zhu, Vignesh Edithal, Le~Zhang, Ilia Blank, and Imran Junejo,
\newblock ``Semantically-aware game image quality assessment,''
\newblock {\em arXiv preprint arXiv:2505.11724}, 2025.

\bibitem{kastryulin2022pytorch}
Sergey Kastryulin, Jamil Zakirov, Denis Prokopenko, and Dmitry~V Dylov,
\newblock ``Pytorch image quality: Metrics for image quality assessment,''
\newblock {\em arXiv preprint arXiv:2208.14818}, 2022.

\bibitem{hammou2025image}
Dounia Hammou, Yancheng Cai, Pavan Madhusudanarao, Christos~G Bampis, and Rafa{\l}~K Mantiuk,
\newblock ``Do image and video quality metrics model low-level human vision?,''
\newblock {\em arXiv preprint arXiv:2503.16264}, 2025.

\bibitem{ma2025survey}
Chengqian Ma, Zhengyi Shi, Zhiqiang Lu, Shenghao Xie, Fei Chao, and Yao Sui,
\newblock ``A survey on image quality assessment: Insights, analysis, and future outlook,''
\newblock {\em arXiv preprint arXiv:2502.08540}, 2025.

\bibitem{koniq10k}
V.~{Hosu}, H.~{Lin}, T.~{Sziranyi}, and D.~{Saupe},
\newblock ``Koniq-10k: An ecologically valid database for deep learning of blind image quality assessment,''
\newblock {\em IEEE Transactions on Image Processing}, vol. 29, pp. 4041--4056, 2020.

\bibitem{li2019quality}
Dingquan Li, Tingting Jiang, and Ming Jiang,
\newblock ``Quality assessment of in-the-wild videos,''
\newblock in {\em Proceedings of the 27th ACM international conference on multimedia}, 2019, pp. 2351--2359.

\bibitem{ghadiyaram2017perceptual}
Deepti Ghadiyaram and Alan~C Bovik,
\newblock ``Perceptual quality prediction on authentically distorted images using a bag of features approach,''
\newblock {\em Journal of vision}, vol. 17, no. 1, pp. 32--32, 2017.

\bibitem{dong2025exploring}
Guanglu Dong, Xiangyu Liao, Mingyang Li, Guihuan Guo, and Chao Ren,
\newblock ``Exploring semantic feature discrimination for perceptual image super-resolution and opinion-unaware no-reference image quality assessment,''
\newblock {\em arXiv preprint arXiv:2503.19295}, 2025.

\bibitem{yang2024no}
Yuxuan Yang, Zhichun Lei, and Changlu Li,
\newblock ``No-reference image quality assessment combining swin-transformer and natural scene statistics,''
\newblock {\em Sensors}, vol. 24, no. 16, pp. 5221, 2024.

\bibitem{song2021ie}
Tianshu Song, Leida Li, Hancheng Zhu, and Jiansheng Qian,
\newblock ``Ie-iqa: Intelligibility enriched generalizable no-reference image quality assessment,''
\newblock {\em Frontiers in Neuroscience}, vol. 15, pp. 739138, 2021.

\bibitem{ghadiyaram2015massive}
Deepti Ghadiyaram and Alan~C Bovik,
\newblock ``Massive online crowdsourced study of subjective and objective picture quality,''
\newblock {\em IEEE Transactions on Image Processing}, vol. 25, no. 1, pp. 372--387, 2015.

\bibitem{ahmed2022biq2021}
Nisar Ahmed and Shahzad Asif,
\newblock ``Biq2021: a large-scale blind image quality assessment database,''
\newblock {\em Journal of Electronic Imaging}, vol. 31, no. 5, pp. 053010--053010, 2022.

\bibitem{babu2023no}
Nithin~C Babu, Vignesh Kannan, and Rajiv Soundararajan,
\newblock ``No reference opinion unaware quality assessment of authentically distorted images,''
\newblock in {\em Proceedings of the IEEE/CVF Winter Conference on Applications of Computer Vision}, 2023, pp. 2459--2468.

\bibitem{ying2021patch}
Zhenqiang Ying, Maniratnam Mandal, Deepti Ghadiyaram, and Alan Bovik,
\newblock ``Patch-vq:'patching up'the video quality problem,''
\newblock in {\em Proceedings of the IEEE/CVF conference on computer vision and pattern recognition}, 2021, pp. 14019--14029.

\bibitem{hosu2017konstanz}
Vlad Hosu, Franz Hahn, Mohsen Jenadeleh, Hanhe Lin, Hui Men, Tam{\'a}s Szir{\'a}nyi, Shujun Li, and Dietmar Saupe,
\newblock ``The konstanz natural video database (konvid-1k),''
\newblock in {\em 2017 Ninth international conference on quality of multimedia experience (QoMEX)}. IEEE, 2017, pp. 1--6.

\bibitem{li2023agiqa}
Chunyi Li, Zicheng Zhang, Haoning Wu, Wei Sun, Xiongkuo Min, Xiaohong Liu, Guangtao Zhai, and Weisi Lin,
\newblock ``Agiqa-3k: An open database for ai-generated image quality assessment,''
\newblock {\em IEEE Transactions on Circuits and Systems for Video Technology}, vol. 34, no. 8, pp. 6833--6846, 2023.

\bibitem{lu2024kvq}
Yiting Lu, Xin Li, Yajing Pei, Kun Yuan, Qizhi Xie, Yunpeng Qu, Ming Sun, Chao Zhou, and Zhibo Chen,
\newblock ``Kvq: Kwai video quality assessment for short-form videos,''
\newblock in {\em Proceedings of the IEEE/CVF Conference on Computer Vision and Pattern Recognition}, 2024, pp. 25963--25973.

\bibitem{hu2025multi}
Bo~Hu, Wei Wang, Chunyi Li, Lihuo He, Leida Li, and Xinbo Gao,
\newblock ``A multi-annotated and multi-modal dataset for wide-angle video quality assessment,''
\newblock {\em arXiv preprint arXiv:2501.12082}, 2025.

\bibitem{ruikar2023nits}
Jayesh Ruikar and Saurabh Chaudhury,
\newblock ``Nits-iqa database: a new image quality assessment database,''
\newblock {\em Sensors}, vol. 23, no. 4, pp. 2279, 2023.

\bibitem{kazemi2024exiqa}
Sepehr Kazemi~Ranjbar and Emad Fatemizadeh,
\newblock ``Exiqa: Explainable image quality assessment using distortion attributes,''
\newblock {\em arXiv e-prints}, pp. arXiv--2409, 2024.

\bibitem{huang2022explainable}
Yipo Huang, Leida Li, Yuzhe Yang, Yaqian Li, and Yandong Guo,
\newblock ``Explainable and generalizable blind image quality assessment via semantic attribute reasoning,''
\newblock {\em IEEE Transactions on Multimedia}, vol. 25, pp. 7672--7685, 2022.

\bibitem{amanova2022explainability}
Narbota Amanova, J{\"o}rg Martin, and Clemens Elster,
\newblock ``Explainability for deep learning in mammography image quality assessment,''
\newblock {\em Machine Learning: Science and Technology}, vol. 3, no. 2, pp. 025015, 2022.

\bibitem{jinjin2020pipal}
Gu~Jinjin, Cai Haoming, Chen Haoyu, Ye~Xiaoxing, Jimmy~S Ren, and Dong Chao,
\newblock ``Pipal: a large-scale image quality assessment dataset for perceptual image restoration,''
\newblock in {\em Computer Vision--ECCV 2020: 16th European Conference, Glasgow, UK, August 23--28, 2020, Proceedings, Part XI 16}. Springer, 2020, pp. 633--651.

\bibitem{yang2022maniqa}
Sidi Yang, Tianhe Wu, Shuwei Shi, Shanshan Lao, Yuan Gong, Mingdeng Cao, Jiahao Wang, and Yujiu Yang,
\newblock ``Maniqa: Multi-dimension attention network for no-reference image quality assessment,''
\newblock in {\em Proceedings of the IEEE/CVF conference on computer vision and pattern recognition}, 2022, pp. 1191--1200.

\bibitem{wang2024large}
Puyi Wang, Wei Sun, Zicheng Zhang, Jun Jia, Yanwei Jiang, Zhichao Zhang, Xiongkuo Min, and Guangtao Zhai,
\newblock ``Large multi-modality model assisted ai-generated image quality assessment,''
\newblock in {\em Proceedings of the 32nd ACM International Conference on Multimedia}, 2024, pp. 7803--7812.

\bibitem{lin2020pea265}
Liqun Lin, Shiqi Yu, Liping Zhou, Weiling Chen, Tiesong Zhao, and Zhou Wang,
\newblock ``Pea265: Perceptual assessment of video compression artifacts,''
\newblock {\em IEEE Transactions on Circuits and Systems for Video Technology}, vol. 30, no. 11, pp. 3898--3910, 2020.

\bibitem{lin2023saliency}
Liqun Lin, Yang Zheng, Weiling Chen, Chengdong Lan, and Tiesong Zhao,
\newblock ``Saliency-aware spatio-temporal artifact detection for compressed video quality assessment,''
\newblock {\em IEEE Signal Processing Letters}, vol. 30, pp. 693--697, 2023.

\bibitem{tang2025clip}
Zhenchen Tang, Zichuan Wang, Bo~Peng, and Jing Dong,
\newblock ``Clip-agiqa: Boosting the performance of ai-generated image quality assessment with clip,''
\newblock in {\em International Conference on Pattern Recognition}. Springer, 2025, pp. 48--61.

\bibitem{qu2024bringing}
Bowen Qu, Haohui Li, and Wei Gao,
\newblock ``Bringing textual prompt to ai-generated image quality assessment,''
\newblock in {\em 2024 IEEE International Conference on Multimedia and Expo (ICME)}. IEEE, 2024, pp. 1--6.

\bibitem{zhu2020metaiqa}
Hancheng Zhu, Leida Li, Jinjian Wu, Weisheng Dong, and Guangming Shi,
\newblock ``Metaiqa: Deep meta-learning for no-reference image quality assessment,''
\newblock in {\em Proceedings of the IEEE/CVF conference on computer vision and pattern recognition}, 2020, pp. 14143--14152.

\bibitem{pan2024sgiqa}
Linpeng Pan, Xiaozhe Zhang, Fengying Xie, Haopeng Zhang, and Yushan Zheng,
\newblock ``Sgiqa: semantic-guided no-reference image quality assessment,''
\newblock {\em IEEE Transactions on Broadcasting}, 2024.

\bibitem{kerkouri2022domain}
Mohamed~Amine Kerkouri, Marouane Tliba, Aladine Chetouani, and Alessandro Bruno,
\newblock ``A domain adaptive deep learning solution for scanpath prediction of paintings,''
\newblock in {\em Proceedings of the 19th International Conference on Content-Based Multimedia Indexing}, 2022, pp. 57--63.

\bibitem{tliba2022self}
Marouane Tliba, Mohamed~Amine Kerkouri, Aladine Chetouani, and Alessandro Bruno,
\newblock ``Self supervised scanpath prediction framework for painting images,''
\newblock in {\em Proceedings of the IEEE/CVF Conference on Computer Vision and Pattern Recognition}, 2022, pp. 1539--1548.

\bibitem{chetouani2020use}
Aladine Chetouani and Leida Li,
\newblock ``On the use of a scanpath predictor and convolutional neural network for blind image quality assessment,''
\newblock {\em Signal Processing: Image Communication}, vol. 89, pp. 115963, 2020.

\bibitem{singh16emotion}
Simranjit Singh, Amrik Singh, and Baljinder Kaur,
\newblock ``Emotion detection through facial expressions: A survey of ai-based methods,''
\newblock {\em IJSAT-International Journal on Science and Technology}, vol. 16, no. 1.

\bibitem{jha2025human}
Bikash~Kumar Jha, Bharat Paudel, Adarsh Mishra, Aabik Maharjan, and Pralhad Chapagain,
\newblock ``Human emotion detection and face recognition system,''
\newblock {\em International Journal on Engineering Technology}, vol. 2, no. 2, pp. 90--97, 2025.

\bibitem{xu2021subjective}
Jun Xu, Weisi Lin, and Leida Zhang,
\newblock ``A subjective study of image quality assessment metrics using crowdsourcing,''
\newblock {\em IEEE Transactions on Image Processing}, vol. 30, pp. 1788--1800, 2021.

\bibitem{li2024survey}
Dongyuan Li, Zhen Wang, Yankai Chen, Renhe Jiang, Weiping Ding, and Manabu Okumura,
\newblock ``A survey on deep active learning: Recent advances and new frontiers,''
\newblock {\em IEEE Transactions on Neural Networks and Learning Systems}, 2024.

\bibitem{ivanjko2019crowdsourcing}
Tomislav Ivanjko,
\newblock ``Crowdsourcing image descriptions using gamification: a comparison between game-generated labels and professional descriptors,''
\newblock in {\em 2019 42nd International Convention on Information and Communication Technology, Electronics and Microelectronics (MIPRO)}. IEEE, 2019, pp. 537--541.

\bibitem{EIJSBROEK2025152603}
Veerle~C. Eijsbroek, Katarina Kjell, H.~Andrew Schwartz, Jan~R. Boehnke, Eiko~I. Fried, Daniel~N. Klein, Peik Gustafsson, Isabelle Augenstein, Patrick~M.M. Bossuyt, and Oscar~N.E. Kjell,
\newblock ``The leading guideline: Reporting standards for expert panel, best-estimate diagnosis, and longitudinal expert all data (lead) methods,''
\newblock {\em Comprehensive Psychiatry}, p. 152603, 2025.

\bibitem{ge2025scalingsyntheticdatacreation}
Tao Ge, Xin Chan, Xiaoyang Wang, Dian Yu, Haitao Mi, and Dong Yu,
\newblock ``Scaling synthetic data creation with 1,000,000,000 personas,'' 2025.

\bibitem{fabbri-etal-2022-qafacteval}
Alexander Fabbri, Chien-Sheng Wu, Wenhao Liu, and Caiming Xiong,
\newblock ``{QAF}act{E}val: Improved {QA}-based factual consistency evaluation for summarization,''
\newblock in {\em Proceedings of the 2022 Conference of the North American Chapter of the Association for Computational Linguistics: Human Language Technologies}, Marine Carpuat, Marie-Catherine de~Marneffe, and Ivan~Vladimir Meza~Ruiz, Eds., Seattle, United States, July 2022, pp. 2587--2601, Association for Computational Linguistics.

\bibitem{Lee2025EvaluatingSR}
Jinu Lee and J.~Hockenmaier,
\newblock ``Evaluating step-by-step reasoning traces: A survey,''
\newblock {\em ArXiv}, vol. abs/2502.12289, 2025.

\bibitem{TID2013}
Nikolay Ponomarenko, Oleg Ieremeiev, Vladimir Lukin, Karen Egiazarian, Lina Jin, Jaakko Astola, Benoit Vozel, Kacem Chehdi, Marco Carli, Federica Battisti, and C.-C.~Jay Kuo,
\newblock ``Color image database tid2013: Peculiarities and preliminary results,''
\newblock in {\em European Workshop on Visual Information Processing (EUVIP)}, 2013, pp. 106--111.

\bibitem{10726865}
Ce~Zhang and Azim Eskandarian,
\newblock ``Image-guided outdoor lidar perception quality assessment for autonomous driving,''
\newblock {\em IEEE Transactions on Intelligent Vehicles}, pp. 1--12, 2024.

\bibitem{10.1145/3705724}
Naeem Ullah, Javed~Ali Khan, Ivanoe De~Falco, and Giovanna Sannino,
\newblock ``Explainable artificial intelligence: Importance, use domains, stages, output shapes, and challenges,''
\newblock {\em ACM Comput. Surv.}, vol. 57, no. 4, Dec. 2024.

\bibitem{Brandenburg2025}
J.~M. Brandenburg, B.~P. M{\"u}ller-Stich, M.~Wagner, et~al.,
\newblock ``Can surgeons trust ai? perspectives on machine learning in surgery and the importance of explainable artificial intelligence (xai),''
\newblock {\em Langenbeck's Archives of Surgery}, vol. 410, pp. 53, 2025.

\bibitem{Balasubramaniam2022}
N.~Balasubramaniam, M.~Kauppinen, K.~Hiekkanen, and S.~Kujala,
\newblock ``Transparency and explainability of ai systems: Ethical guidelines in practice,''
\newblock in {\em Requirements Engineering: Foundation for Software Quality}, Vincenzo Gervasi and Andreas Vogelsang, Eds., vol. 13216 of {\em Lecture Notes in Computer Science}. Springer, Cham, 2022.

\bibitem{10.1145/3617233.3617234}
Aymen Sekhri, Mohamed~A Kerkouri, Aladine Chetouani, Marouane Tliba, Yassine Nasser, Rachid Jennane, and Alessandro Bruno,
\newblock ``Automatic diagnosis of knee osteoarthritis severity using swin transformer,''
\newblock in {\em Proceedings of the 20th International Conference on Content-Based Multimedia Indexing}, New York, NY, USA, 2023, CBMI '23, p. 41–47, Association for Computing Machinery.

\bibitem{EngelHermann2024}
P.~Engel-Hermann and A.~Skulmowski,
\newblock ``Appealing, but misleading: a warning against a naive ai realism,''
\newblock {\em AI Ethics}, 2024.

\bibitem{Masoudnia2014}
Saeed Masoudnia and Reza Ebrahimpour,
\newblock ``Mixture of experts: a literature survey,''
\newblock {\em Artificial Intelligence Review}, vol. 42, pp. 275--293, 2014.

\bibitem{Hu2022}
Edward~J. Hu, Yelong Shen, Phil Wallis, Zeyuan Allen-Zhu, Yuanzhi Li, Shawn Wang, and Weizhu Chen,
\newblock ``Lora: Low-rank adaptation of large language models,''
\newblock in {\em Proceedings of the International Conference on Learning Representations (ICLR)}, 2022, vol.~1, p.~3.

\end{thebibliography}
}

\end{document}